\documentclass{hld2024} 

\title[Interpolated-MLPs: Controllable Inductive Bias]{Interpolated-MLPs: Controllable Inductive Bias}

\hldauthor{%
\Name{Sean Wu}\textsuperscript{*}  \Email{seanwu@ethz.ch}\\
\Name{Jordan Hong}\textsuperscript{*}  \Email{johong@ethz.ch}\\
\Name{Keyu Bai}\textsuperscript{*}  \Email{keybai@ethz.ch}\\
\Name{Gregor Bachmann} \Email{gregorb@ethz.ch}\\
\addr {ETH Zurich, Switzerland}
}

\usepackage{booktabs} 
\usepackage{floatrow}

\begin{document}

\maketitle
\def\thefootnote{*}\footnotetext{Equal contribution.}
\vspace{-1.4cm}
\begin{abstract}
Due to their weak inductive bias, Multi-Layer Perceptrons (MLPs) have subpar performance at low-compute levels compared to standard architectures such as convolution-based networks (CNN).
Recent work, however, has shown that the performance gap drastically reduces as the amount of compute is increased without changing the amount of inductive bias \cite{bachmann2023scaling}.
In this work, we study the converse: in the low-compute regime, how does the incremental increase of inductive bias affect performance?
To quantify inductive bias, we propose a ``soft MLP'' approach, which we coin \textit{Interpolated MLP} (I-MLP). 
We control the amount of inductive bias in the standard MLP by introducing a novel algorithm based on interpolation between fixed weights from a \textit{prior model} with high inductive bias.
We showcase our method using various \textit{prior models}, including CNNs and the MLP-Mixer architecture.
This interpolation scheme allows fractional control of inductive bias, which may be attractive when full inductive bias is not desired (e.g. in the mid-compute regime).
We find experimentally that for Vision Tasks in the low-compute regime, there is a continuous and two-sided logarithmic relationship between inductive bias and performance when using CNN and MLP-Mixer prior models.
\end{abstract}

\vspace{-0.25cm}
\section{Introduction} \label{intro}
MLPs provide a rich theoretical deep learning framework due to their mathematical simplicity despite inferior practical performance.
For vision tasks, modern models based on the Convolutional Neural Network (CNN) \cite{krizhevsky2012imagenet} \cite{he2015deep} and Vision Transformer (ViT) \cite{dosovitskiy2021image} have demonstrated better performance.
A recent hypothesis claims that CNN and ViT are superior to MLPs because inductive bias is detrimental at high-compute scales \cite{dosovitskiy2021image}. \\[2mm]
The work of Bachmann et al. \cite{bachmann2023scaling} recently strengthened this hypothesis and showed that with sufficient parameter count and pre-training, even simple MLPs can achieve strong performance on vision tasks.
We aim to continue this line of research and ask: i) at a low-compute scale, what is the relationship (e.g. linear, logarithmic, exponential) between the amount of inductive bias and performance? ii) since there is a balance of tension (inductive bias is useful in low-compute but not desired in high-compute), can we parameterize (and optimize over) the fractional amount of inductive bias, particularly in the mid-compute regime?\\[2mm]
In this study, we introduce inductive bias to MLPs by methods of \textit{locality injection}, first proposed by Ding et al. \cite{ding2022repmlpnet}, where they converted a convolution kernel $F$ into a fully-connected layer $\mathbf{W_F}$, then merged (adding) $\mathbf{W_F}$ to a parallel fully-connected layer $\mathbf{W}$.
The inductive bias is drawn from the convolutional kernel, which we will refer to as the \textit{prior model}.\\[2mm]
In this study, we extend the locality injection approach in two directions.
First, instead of adding the prior model to the MLP layer, we interpolate using a parameter \(\alpha\) to fractionally control the amount of added inductive bias.
Second, in addition to a CNN, we extend the interpolation to another prior model: the MLP-Mixer \cite{tolstikhin2021mlpmixer}.
For both prior models, we i) investigate the trade-off between the amount of inductive bias \(\alpha\) and performance, and ii) provide architectures and training methods that are performant under compute constraints.

\vspace{-0.25cm}
\section{Background} \label{sec:background}
\paragraph{Notation.} We consider an arbitrary layer with tensor input \(\mathbf{x} \in \mathbb{R}^{c_{in} \times h_{in} \times w_{in}}\) and output \(\mathbf{y} \in \mathbb{R}^{c_{out} \times h_{out} \times w_{out}}\ \), where \(c_{in}, h_{in}, w_{in}\) denote the channel, height, and width of the input, \(c_{out}, h_{out}, w_{out}\) denote the same for the output. For MLP layers, the tensor input and output are vectorized into a single dimension: \(vec(\mathbf{x}) \in \mathbb{R}^{c_{in}h_{in}w_{in}}\) and \(vec(\mathbf{y} )\in \mathbb{R}^{c_{out}h_{out}w_{out}}\).
The equivalent fully-connected layer representation of a \textit{prior model} is denoted as \(\mathbf{W_P} \in \mathbb{R}^{c_{out}h_{out}w_{out} \times c_{in}h_{in}w_{in}}\).
\paragraph{Standard MLP.} As a starting point, we consider the standard MLP (S-MLP) as in \cite{bachmann2023scaling} with an input $\mathbf{z}$ of general dimension $M_{in}$, weight matrix $\mathbf{W} \in \mathbb{R}^{M_{out}\times M_{in}}$, nonlinear activation $\sigma$, and Layer Normalization $\operatorname{LN}$ :
\begin{equation}
    \operatorname{Block}(\mathbf{z}) = \sigma\left(  \operatorname{LN}(\mathbf{z}\mathbf{W}^T)\right).
\end{equation}
For vision tasks, the tensor input and output are vectorized, i.e. \(\mathbf{z} = vec(\mathbf{x})\), \(M_{in} = c_{in}h_{in}w_{in}\), and \(M_{out} = c_{out}h_{out}w_{out}\). 
    \begin{equation}
        vec(\mathbf{y}) = vec(\mathbf{x}) \mathbf{W}^T .
        \label{eq:mlp-layer}
    \end{equation}
The flattening procedure removes all locality of an image tensor. As such, the MLP does not possess any inductive bias for vision tasks. However, such locality is particularly relevant to images.
Next, we introduce the two prior models: CNN and MLP-Mixer.
\paragraph{CNN.} Convolution can be viewed as a special case of an MLP layer, in which a special weight matrix $\mathbf{W_F}$ is structured by being sparse and having shared weights, which leads to spatially localized learning and introduces inductive bias useful for vision tasks \cite{bachmann2023scaling}.
We use  $F$ to denote the convolution kernel and express the output tensor as:
\begin{equation}
    \mathbf{x}^{(out)} = F * \mathbf{x}^{(in)}.
    \label{eq:cnn-layer}
\end{equation}
Equation \eqref{eq:cnn-layer} can be expressed using the MLP formulation with vectorized input and output.
\begin{equation}
    vec(\mathbf{x}^{(out)})= vec(F * \mathbf{x}^{(in)}) = vec(\mathbf{x}^{(in)}) \mathbf{W_F}^T,
    \label{eq:mlp-cnn-equiv-layer}
\end{equation}
where \(\mathbf{W_F} \in \mathbb{R}^{c_{out}h_{out}w_{out} \times c_{in}h_{in}w_{in}} \) is the equivalent fully-connected layer obtained from \(F\).
Here, \(\mathbf{W_P} = \mathbf{W_F}\).
Appendix \ref{app:cnn-conversion} shows the explicit construction from \(F\) to \(\mathbf{W_F}\).

\paragraph{MLP-Mixer.} 
We identify three sources of inductive bias in the MLP-Mixer, all of which we model as a special case of a fully-connected layer: i) Patchifying and per-patch linear embeddings, ii) Token-mixing with weight sharing, iii) Channel-mixing with weight sharing. These inductive biases allow the MLP-Mixer to attain competitive results and perform comparably to the current state-of-the-art ViT \cite{tolstikhin2021mlpmixer}.
Collectively, the above operations can be expressed as two successive operations: a linear operation \(\mathbf{L} \in \mathbb{R}^{c_{in}h_{in}w_{in} \times c_{in}h_{in}w_{in}}\), and a Toeplitz matrix $\Tilde{\mathbf{W}}\in \mathbb{R}^{c_{out}h_{out}w_{out} \times c_{in}h_{in}w_{in}}$ acting on the vectorized input $vec(\mathbf{x})$.
The linear operation \(\mathbf{L}\) can be either a patchifying matrix \(\mathbf{P}\), a transpose matrix \(\mathbf{T}\), or an identity matrix \(\mathbf{I}\) in the degenerate case.
\begin{equation}
    vec(\mathbf{y}) =  vec(\mathbf{x}) \mathbf{L}^T \Tilde{\mathbf{W}}^T
    \label{eq:mlp-mlp-mixer-equiv-layer}
\end{equation}
The fully-connected equivalent matrix is given by \(\mathbf{W_P} =  \Tilde{\mathbf{W}} \mathbf{L} \) and derived in Appendix \ref{app:mlp-mixer-conversion} .
\begin{figure}[htpb]
    \centering
    \includegraphics[width=\textwidth]{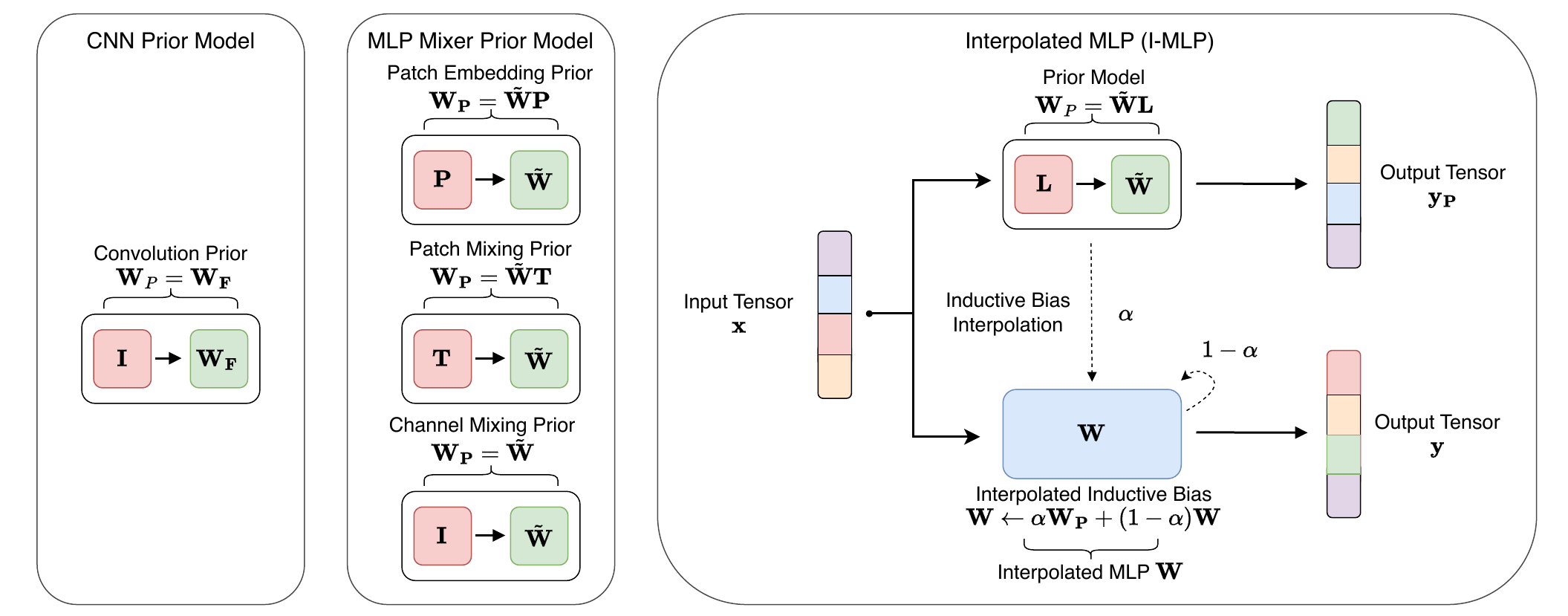}
    \caption{Graphic description of the Interpolated MLP (left), the CNN prior (middle) and the MLP-Mixer prior (right).}
    \label{fig:MLP-Mixer}
\end{figure}

\vspace{-0.25cm}
\section{Interpolation Training Method}
\label{sec:interp_train_method}
Algorithm \ref{alg:training-method} outlines the training method for our proposed Interpolated MLP (I-MLP), which interpolates with a prior model layer in every epoch. For a multi-layer network, the update rule in Algorithm \ref{alg:training-method} is applied independently to all layers.
In this paper, we study the interpolation for two prior models: CNN and MLP-Mixer. 
We obtain $\mathbf{W_P}$ using Equation \ref{eq:mlp-cnn-equiv-layer} for CNNs and Equation \ref{eq:mlp-mlp-mixer-equiv-layer} for MLP-Mixers.

\begin{algorithm2e}[h!]
    \DontPrintSemicolon
    \SetKwInOut{Input}{Initialize}
    \Input{I-MLP weight matrix $\mathbf{W}$ (fully-connected) and prior model $P$}
    Training:\;
    \ForEach{epoch} {
        Independently train $\mathbf{W}$ and $P$ (forward and back propagation).\;
        Interpolate $\mathbf{W}$: $\mathbf{W} \gets (1- \alpha) \mathbf{W} + \alpha \mathbf{W_P}$, where $\mathbf{W_P}$ is the converted fully-connected layer from the prior model $P$.\;
    }
    \label{alg:training-method}
    \caption{Interpolated MLP (I-MLP) training method}
\end{algorithm2e}

When $\alpha = 1$, the interpolated $\mathbf{W}$ is equivalent to the prior model $P$. When $\alpha =0$, the interpolated $\mathbf{W}$ receives no inductive bias and is just a pure fully-connected layer with no structural constraints.
By adjusting $\alpha$, we can control the amount of inductive bias introduced into the MLP.

\vspace{-0.25cm}
\section{Experiments}
We investigate the impact of interpolated inductive bias with several experiments.
For the two prior models (CNN and MLP-Mixer), we design parallel I-MLP structures.
We then train the pair of models (prior model, I-MLP) according to Section \ref{sec:interp_train_method} and vary the value of \(\alpha\) between 0 and 1.
Our Interpolated-MLP model follows the Standard MLP (S-MLP) architecture described by Bachmann et al. \cite{bachmann2023scaling}.
The experimental setup can be found in Appendix \ref{app:experimental-setup}.

\vspace{-0.1cm}
\subsection{Interpolation with Varying $\alpha$}
\begin{figure}
\centering
\subfigure[Prior model: CNN]{
    \includegraphics[width=0.45\textwidth]{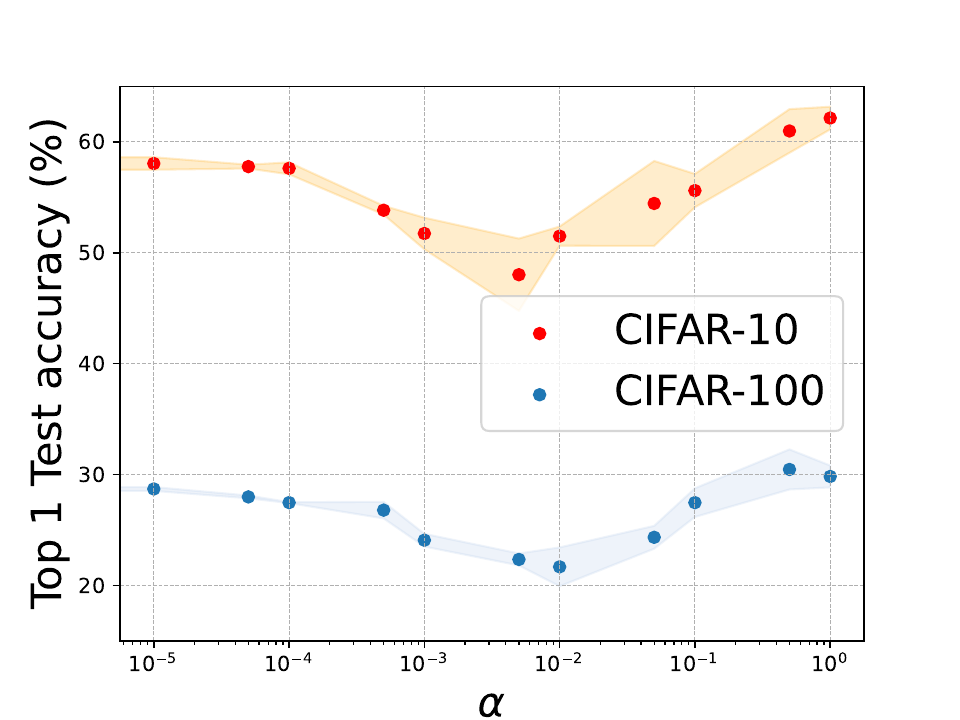}
}
\subfigure[Prior model: MLP-Mixer]{
\includegraphics[width=0.45\textwidth]{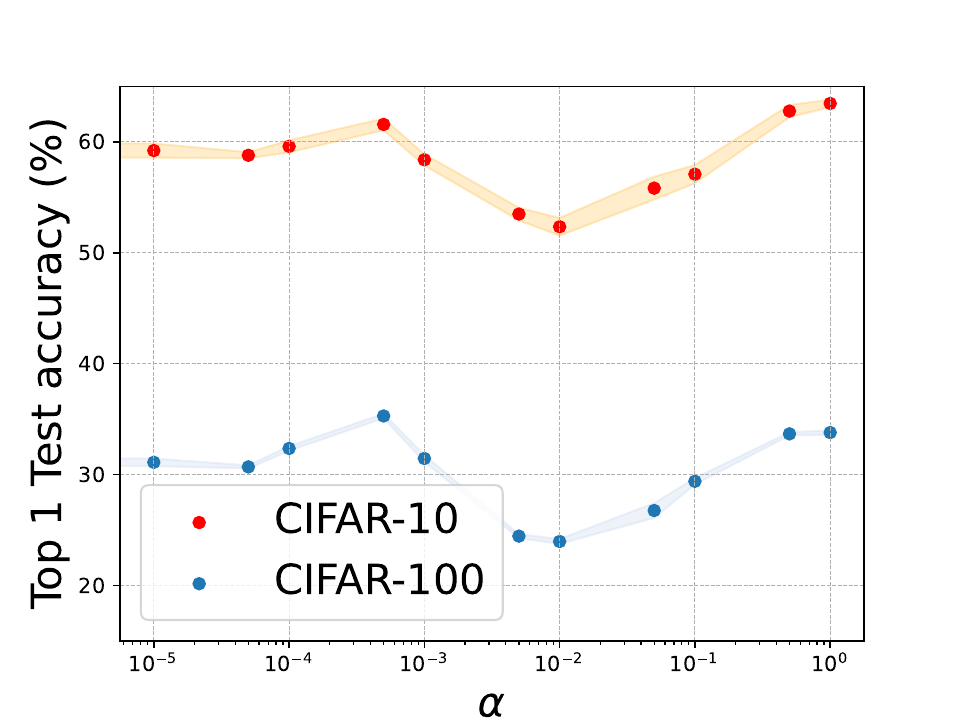}
}
\vspace{-10pt}
\caption{Test accuracy of I-MLP with varying $\alpha$ (with data augmentation).}
\label{fig:imlp-cnn-mixer-test_acc_vs_alpha}
\end{figure}

To interpolate between a CNN and an MLP, we use the Standard CNN (S-CNN) based on AlexNet \cite{krizhevsky2012imagenet} as our prior model.
The CNN has six convolutional layers, each with an output dimension of \(d_{CNN} = o \, h_{out} \, w_{out}\). 
The I-MLP (with CNN prior) has six fully-connected layers, each with a width of \(d_{MLP} = 1024\). 
To ensure proper interpolation, we meticulously match the layer dimensions between the pair; at each layer, \(d_{CNN} = d_{MLP}\). The detailed method and exact layer dimensions are outlined in Appendix \ref{app:I-MLP-arch-detail-CNN}.\\[2mm]
Similarly, for interpolating between an MLP-Mixer and an MLP, we use the original MLP-Mixer from \cite{tolstikhin2021mlpmixer} as our prior model, with specific dimensions outlined in the appendix.
We design the I-MLP (with MLP-Mixer prior) by replacing the three sources of inductive bias with fully-connected layers. 
The interpolation method is detailed in Sections \ref{app:linear_transform} and \ref{app:weight_sharing}, and the architectures of the two models are described in Appendix \ref{app:experimental-setup} (Table \ref{tab:mlp_mixer_model_arch}).\\[2mm]
We vary the interpolation weight $\alpha$ and plot the corresponding top 1 test accuracies on a semilogarithmic plot in Figure \ref{fig:imlp-cnn-mixer-test_acc_vs_alpha} with a CNN prior (left) and MLP-Mixer prior (right).
In Figure \ref{fig:imlp-cnn-mixer-test_acc_vs_alpha} (left), we observe a minimum at around $\alpha = 5 \times 10^{-3}$, with either side of the minimum exhibiting logarithmic behavior. 
We hypothesize that the S-MLP and S-CNN models are converging to different local minima in the loss landscape; when we interpolate between them, there exists an $\alpha$ such that the two minima interfere with equal strength, resulting in poor performance. 
Decreasing $\alpha$ below $5 \times 10^{-3}$ improves performance logarithmically, approaching S-MLP levels. 
Similarly, increasing $\alpha$ above $5 \times 10^{-3}$ improves performance logarithmically, approaching S-CNN levels.
In Figure \ref{fig:imlp-cnn-mixer-test_acc_vs_alpha} (right), the same logarithmic behavior is observed for the I-MLP with MLP-Mixer prior.
However, we observe both a minimum at \(\alpha = 10^{-2}\) and a (local) maximum at \(\alpha = 5 \times 10^{-4}\).
We suspect that this is because, in MLP-Mixer, we are simultaneously interpolating three different sources of inductive bias (patchifying, token-mixing, and channel-mixing) as outlined in Section \ref{sec:background}.
Each inductive bias exhibits a minimum at a different \(\alpha\).
The final result is the superposition of multiple different V-shapes, with potential constructive and/or destructive interference.\\[2mm]
At the endpoints, the I-MLP performance approaches S-MLP at $\alpha = 0$ and the prior model (S-CNN or MLP-Mixer) at $\alpha = 1$.
\vspace{-2mm}

\begin{table}[!h]
    
    \centering
    \begin{small}
        \scriptsize
        \centering
        \subtable[I-MLP (with CNN prior)][]{
            \begin{tabular}{ccc}
                \toprule
                Model & CIFAR-10 & CIFAR-100 \\
                \midrule
                I-MLP ($\alpha=0.0$) & 58.36 $\pm$ 0.44 & 29.06 $\pm$ 0.21 \\
                I-MLP ($\alpha=0.01$) & 51.50 $\pm$ 0.86 & 21.67 $\pm$ 1.75 \\
                I-MLP ($\alpha=1.0$) & 62.14 $\pm$ 1.02 & 29.82 $\pm$ 0.99 \\
                S-CNN & 62.60 $\pm$ 1.20 & 32.15 $\pm$ 0.57 \\
                \bottomrule
            \end{tabular}
        }
        \subtable[I-MLP (with MLP-Mixer prior)][]{
            \begin{tabular}{ccc}
                \toprule
                Model & CIFAR-10 & CIFAR-100 \\
                \midrule
                I-MLP ($\alpha=0.0$) & 59.35 $\pm$ 0.22 & 31.00 $\pm$ 0.21 \\
                I-MLP ($\alpha=0.01$) & 52.34 $\pm$ 0.80 & 23.96 $\pm$ 0.22 \\
                I-MLP ($\alpha=1.0$) & 63.51 $\pm$ 0.11 & 33.89 $\pm$ 0.20 \\
                MLP-Mixer & 63.42 $\pm$ 0.48 & 33.71 $\pm$ 0.13 \\
                \bottomrule
            \end{tabular}
        }
    \end{small}
    \caption{CIFAR-10 and CIFAR-100 test dataset top 1 accuracy scores (mean $\pm$ standard deviation). We include \(\alpha = 1.0\) to emphasize the equivalence of I-MLP-CNN with S-CNN and I-MLP-Mixer with MLP-Mixer when \(\alpha = 1\). Numerical results confirm this.}
    \label{tab:imlp_dataset_augmentation}
\end{table}

\vspace{-1cm}
\subsection{Additional Experiments: Test-time Only Interpolation, Interpolation Weight Decay, and First-layer Only Interpolation}
We perform three additional experiments with different interpolation strategies. In Appendix \ref{app:test-time-only-interp}, we compare interpolation during training with test-time only interpolation.
We separately train the MLP and prior models, and only use the interpolation weight once before test and inference time.
The results in Table \ref{tab:nonconstant_interp} show that test-time only interpolation performs significantly worse. \\[2mm]
In Appendix \ref{app:interp-weight-decay}, we experiment with non-constant interpolation weights during training with $\alpha[t]$ as a function of the epoch $t$.
We define a decaying interpolation weight schedule of $\alpha[t] = a(1-\frac{t}{t_{max}})^k$, where $a$ represents the initial interpolation weight and $k$ controls the decay rate.
We observe in Figure \ref{fig:interpolation_weight_decay} that interpolation with a constant weight $k=0$ performs similarly to strong weight decay $k>4$, with linear decay $k=1$ performing significantly worse.\\[2mm]
In Appendix \ref{app:first-layer-interp}, we experiment with more time- and space-efficient interpolation strategies by limiting the number of interpolation operations.
We constrain the number of weight matrix parameters being interpolated at each epoch and compare interpolating only a wider first layer versus interpolating multiple narrower layers.
For this fixed interpolation parameter budget, we observe that a wide first layer interpolation with a CNN prior is better as shown in Table \ref{tab:result_comparison}.
This empirical finding is important because we can improve training computational efficiency and potentially extend our results to larger datasets than CIFAR-10 and CIFAR-100.

\vspace{-0.25cm}
\section{Summary}
In this work, we explore novel training techniques that allow one to continuously increase or decrease the inductive bias present in a  plain MLP.
Through an interpolation strategy with more structured priors, we study the relationship between MLP performance and fractional inductive bias at low-compute scales.
Our experiments strengthen previous results on the role of inductive bias; we believe that our novel training technique can be useful for more controlled analyses in this line of work. For example, in the mid-compute regime, a fractional amount of inductive bias may be the optimal design.
Future work could explore more involved interpolation strategies and advanced data augmentation pipelines.
Experimenting with larger compute scales and investigating the role of inductive bias in such a regime makes for exciting future work.

\bibliography{main}

\appendix

\section{Construction of \(\mathbf{W_p}\) from a CNN layer} \label{app:cnn-conversion}
The conversion of a 2-D convolution kernel $F$ into a fully connected layer $\mathbf{W_F}$ was derived by \cite[Section 3]{ding2022repmlpnet}.
We adapt the formula here to allow for different input and output dimensions, i.e. when $h_{out} \neq h_{in}$ and $w_{out} \neq w_{in}$:
\begin{equation}
    \mathbf{W_F} = \text{reshape} ( F*\mathbf{x}^{(I)}, (ch_{in}w_{in}, oh_{out}w_{out}))^T,
\end{equation}
where
\begin{equation}
    \mathbf{x}^{(I)} = \text{reshape} (\mathbf{I}, (ch_{in}w_{in}, c, h_{in}, w_{in})),
\end{equation}
and $\mathbf{I}$ is the identity matrix with dimensions $(ch_{in}w_{in}, ch_{in}w_{in})$.
The function $\text{reshape} (\mathbf{x}, dim)$ reshapes a tensor $\mathbf{x}$ into dimension $dim$ in-order.

\section{Construction of \(\mathbf{W_p}\) from MLP-Mixer} \label{app:mlp-mixer-conversion}
For the MLP-Mixer, we require the equivalent fully-connected layer representation \(\mathbf{W_P} =  \Tilde{\mathbf{W}} \mathbf{L} \), where \(\mathbf{L} = \mathbf{P}\) or \(\mathbf{L} = \mathbf{T}\) for the patchify and transpose operations, and \(\Tilde{\mathbf{W}}\) corresponds to the expanded Toeplitz matrix for weight sharing.
In the following subsections, we show the constructions for \(\mathbf{L} = \mathbf{T}\) (Appendix \ref{app:patch_transform}), \(\mathbf{L} = \mathbf{T}\) (Appendix \ref{app:transpose_transform}), and \(\Tilde{\mathbf{W}}\) (Appendix \ref{app:weight_sharing}).

\subsection{Linear operation matrix \(\mathbf{L}\)} \label{app:linear_transform}

\subsubsection{Patchifying matrix $\mathbf{P}$} \label{app:patch_transform}
We first show the construction when \(\mathbf{L}=\mathbf{P}\) is the patchifying matrix for tensor \(\mathbf{x}\). 
For notational simplicity, we assume a single channel, i.e. \(\mathbf{x}\in \mathbb{R}^{H \times W}\).
We denote:
\begin{itemize}
    \item \((H, W)\): dimensions of the input tensor.
    \item \((h, w)\): Number of patches along height and width. \( h = \frac{H}{P} \) and \( w = \frac{W}{P} \).
    \item \((P, P)\): dimensions of each patch.
    \item \((R, C)\): patch index (denoting the order of one patch), \( R \in \{0, 1, \ldots, h-1\} \) and \( C \in \{0, 1, \ldots, w-1\} \).
    \item \((k, l)\) : (absolute) pixel index within the tensor \(\mathbf{x}\).
    \item \((r, c)\): (relative) pixel index (denoting the order of pixels within the patch). \( r, c \in \{0, 1, \ldots, P-1\} \).
\end{itemize}
The \((k,l)\)-th pixel in tensor \(\mathbf{x}\) takes the following position in $vec(\mathbf{x})$.
\begin{equation}
    \label{eq:func_kl_to_idx}
    idx(k,l) := (k\cdot W + l)
\end{equation}
For a given patch \((R, C)\) and relative pixel within the patch \((r, c)\), the top-left corner of patch \((R, C)\) in the original image is at:
\begin{equation}
    \label{eq:RC_to_k0l0}
    (k_0, l_0) = (R \cdot P, C \cdot P).
\end{equation}
For each element within the patch, the corresponding index in tensor \(\mathbf{x}\) is:
\begin{equation}
    \label{eq:rc_to_kl}
    (k, l) = (k_0 + r, l_0 + c)
\end{equation}
Substituting \eqref{eq:rc_to_kl} to \eqref{eq:func_kl_to_idx} gives the absolute index in $\mathbf{x}$:
\begin{equation}
    idx(r,c) = (k_0 + r) \cdot W + (l_0 + c).
\end{equation}
Next, we substitute \((k_0, l_0)\) using \eqref{eq:RC_to_k0l0}:
\begin{equation}
    \label{RCrc_to_idx}
    idx(R,C,r,c) = (R\cdot P+ r) \cdot W + (C\cdot P + c).
\end{equation}

\paragraph{Construction of P.} 
Observe that \(\mathbf{P}_{ij} = 1\) denotes ``\(x_j\) should be permuted to position \(i\)".

For fixed \(i\), we first calculate the patch number and \((R,C)\), as well as the relative pixel number and \((r,c)\). 
The patch number is: 
\begin{equation}
    \left\lfloor \frac{i}{P^2} \right\rfloor,
\end{equation}
with patch index:  
\begin{equation}
    \label{eq:RC}
    (R,C) = \left( \left\lfloor \left(\frac{i}{P^2}\right)\frac{1}{w} \right\rfloor, \left(\frac{i}{P^2}\right) \bmod w \right)
\end{equation}
Within the patch \((R,C)\), the pixel has pixel number: 
\begin{equation}
    i \bmod P^2,
\end{equation}
and relative pixel index:
\begin{equation}
    \label{eq:rc}
    (r,c) = \left( \left\lfloor \left(i\bmod P^2\right)\frac{1}{P} \right\rfloor, (i \bmod P^2) \bmod P\right).
\end{equation}
Finally, we use \((R,C)\) and \((r,c)\) and calculate the source index \(j\) of input tensor \(\mathbf{x}\):
\begin{equation}
\mathbf{P}_{ij} = 
\begin{cases}
1 & \text{if } j = (R\cdot P + r) \cdot W + (C\cdot P + c), \\
0 & \text{otherwise},
\end{cases}
\end{equation}
where \((R,C)\) and  \((r,c)\) follow from \eqref{eq:RC} and \eqref{eq:rc}.

\subsubsection{Transpose matrix \(\mathbf{T}\)} \label{app:transpose_transform}
We now show the construction of the transpose matrix \(\mathbf{L}=\mathbf{T}\) to transpose a single channel tensor \(\mathbf{x}\in \mathbb{R}^{H \times W}\) to \(\mathbf{x^T}\in \mathbb{R}^{W \times H}\).

We adapt the in-place transpose permutation algorithm from \cite{cate1977algorithm}, where $\pi(k)$ indicates the original index into a 2D matrix stored row-wise and 
$k$ indicates the new index after in-place transposition.
\begin{equation}
    \pi(k) = 
    \begin{cases}
        Wk \bmod{HW-1}  & \text{if } k \neq HW-1, \\
        HW-1            & \text{if } k=HW-1.
    \end{cases}
\end{equation}

Note that \(\mathbf{T}_{ij} = 1\) denotes ``\(x_j\) should be permuted to position \(i\)".
The desired transpose transformation is then:

\begin{equation}
    \mathbf{T}_{ij} = 
    \begin{cases}
        1 & \text{if } j = \pi(i), \\
        0            & \text{otherwise}, 
    \end{cases}
\end{equation}

\subsection{Equivalent fully connected layer for weight sharing} \label{app:weight_sharing}
Consider an input tensor $\mathbf{X} \in \mathbb{R}^{R \times C}$.
We can interpret this input tensor as a set of row vector inputs $\mathbf{x_r} \in \mathbb{R}^{C}$.
The MLP-Mixer, for example, uses a set of row vectors $\mathbf{x_r}$ in its Channel Mixer where each $\mathbf{x_r}$ represents a patch.

We can apply the same MLP layer to each of the row vectors.
We denote this as an MLP with shared weights because the same weights are used for each row vector.
For simplicity, we express the shared weight matrix formulation using weight matrices where the input and output dimensions are the same.
Note, however, that the linear patch embeddings in MLP-Mixer do not use isotropic weight matrices.

The shared weight $\mathbf{w_r} \in \mathbb{R}^{C\times C}$ acts on each row of the input tensor $\mathbf{X} \in \mathbb{R}^{R \times C}$ and produces an output tensor $\mathbf{Y} \in \mathbb{R}^{R \times C}$. For the $i$-th row,
\begin{equation}
    \mathbf{y_i} = \mathbf{x_i} \mathbf{w_r}^T, \quad i = 1, \cdots, R
    \label{eq:row-major}
\end{equation}
where $x_i \in \mathbb{R}^C$ and $y_i \in \mathbb{R}^C$ are the $i$-th row of $\mathbf{X}$ and $\mathbf{Y}$ respectively.
Equation \eqref{eq:row-major} can be alternatively expressed as a Toeplitz matrix $\mathbf{W} \in \mathbb{R}^{RC \times RC}$ acting on the vectorized form of $\mathbf{X}$ and $\mathbf{Y}$:
\begin{equation}
    vec(\mathbf{Y}) = vec(\mathbf{X}) \mathbf{W}^T,
\end{equation}
where the shared linear layer layer $\mathbf{w}$ is repeated on the diagonal of $\mathbf{W}$.
\begin{equation}
    [ \mathbf{y_1}, \mathbf{y_2}, \dots, \mathbf{y_R} ] =
    [ \mathbf{x_1}, \mathbf{x_2}, \dots, \mathbf{x_R} ]
        \begin{bmatrix}
        \begin{array}{ccccc}
         \mathbf{w_r} & 0 & \cdots & 0 \\
         0 & \mathbf{w_r} & \cdots & 0 \\
         \vdots & \vdots &   \ddots   & \vdots         \\
          0 &   0  & \cdots  & \mathbf{w_r}
        \end{array}
    \end{bmatrix}^T
\end{equation}
Therefore, the explicit weight matrix is $\mathbf{W} = diag(\mathbf{w_r})$.

\section{I-MLP Architecture Details - CNN} \label{app:I-MLP-arch-detail-CNN}
We design an Interpolated MLP (I-MLP) that interpolates between the S-MLP and S-CNN.
We summarize the S-MLP, I-MLP, and S-CNN architectures in Table \ref{tab:model_arch} for batch size $n$ and image shape $32 \times 32 \times 3$.
To allow for interpolation, we require that in each layer, the dimensions are consistent across all three models, i.e. $d_{MLP} = d_{CNN}$.
Since CNNs require multiple channels to be performant, we increase the output channel $o$ of each layer in an encoder-like fashion.
To counteract the increase in channel dimensions, we decrease $h_{out},w_{out}$ by using a kernel size $k \times k = 3 \times 3$, stride $s=2$, and padding $p=1$ to satisfy the $d_{CNN} = d_{MLP} = 1024$ constraint.
The layer output height $h_{out}$ and width $w_{out}$ are governed by $h_{out} = \bigg(\frac{h_{in} - k + 2p}{s}\bigg) + 1$ and $w_{out} = \bigg(\frac{w_{in} - k + 2p}{s}\bigg) + 1$.
For all three architectures, the first layer serves as an embedding layer to convert images with 3 RGB channels to 1 channel and the final layer is a linear classifier.
Following \cite{bachmann2023scaling}, we use Layer Norm for normalization and GELU as the activation function.

\begin{table}[h]
    \vskip 0.15in
    \centering
    \begin{small}
    \centering
    \begin{tabular}{llll}
        \toprule
        MLP Layer & CNN Layer & MLP Output Shape & CNN Output Shape \\
        \midrule
        \textbf{Layer-1} & Convolution 2D & [n, 1024] & [n, 1, 32, 32] \\
        \midrule
        \textbf{Layer-2} & Convolution 2D & [n, 1024] & [n, 4, 16, 16] \\
        \midrule
        \textbf{Layer-3} & Convolution 2D & [n, 1024] & [n, 16, 8, 8] \\
        \midrule
        \textbf{Layer-4} & Convolution 2D & [n, 1024] & [n, 64, 4, 4] \\
        \midrule
        \textbf{Layer-5} & Convolution 2D & [n, 1024] & [n, 256, 2, 2] \\
        \midrule
        \textbf{Layer-6} & Convolution 2D & [n, 1024] & [n, 256, 2, 2] \\
        \midrule
        Linear-7 & Classification Head & [n, 10] & [n, 10] \\
        \bottomrule
    \end{tabular}
    \end{small}
    \caption{Model architectures for MLP (S-MLP and I-MLP) and CNN (S-CNN) networks used. The fully connected and convolutional layers used for interpolation are bolded. For both models, the features from Layers-1 and Layers-5 are normalized (through a layer norm) then activated using GELU. The S-MLP and I-MLP models have 8,405,002 parameters, while the S-CNN model uses 757,982 parameters.}
    \label{tab:model_arch}
\end{table}

\section{I-MLP Architecture Details - MLP-Mixer} \label{app:I-MLP-arch-detail-MLP-mixer}
We design an Interpolated MLP (I-MLP) that interpolates between the S-MLP and the MLP-Mixer. 
We summarize the S-MLP, I-MLP, and MLP-Mixer architectures in Table \ref{tab:mlp_mixer_model_arch} for batch size $n$ and image shape $32 \times 32 \times 3$.

The MLP-Mixer consists of repeated groups of layers called Mixer Layers, with the MLP-Mixer depth indicated the number of Mixer Layers.
Each Mixer Layer operates on a 2D tensor $\mathbf{x} \in \mathbb{R}^{S \times C}$, where $S$ indicates the number of patches and $C$ indicates the number of channels after performing linear patch embedding.
Within each Mixer Layer are two 2-layer MLPs with shared weights called the Token Mixer and the Channel Mixer.
First, the Token Mixer acts on the patch dimension; it applies its MLP layers with shared weights to each column vector $\mathbf{x_c} \in \mathbb{R}^S$.
Then, the Channel Mixer acts on the channel dimension: it applies its MLP layers with shared weights to each row vector $\mathbf{x_r} \in \mathbb{R}^C$.
To implement this behavior, we apply a transpose linear transformation $\mathbf{L} = \mathbf{T}$ to the input and output of the Token Mixer.
This approach allows us to use the same matrix $\Tilde{\mathbf{W}}$ formulation for weight sharing along rows.

As with the I-MLP with CNN prior, we require for interpolation that in each layer, the dimensions are consistent across all three models, i.e. $d_{MLP} = d_{Mixer}$.
We use an $8 \times 8$ patch size, $S = \frac{HW}{p^2} = 16$ patches, $C=128$ hidden channels, and an MLP-Mixer depth of 2.
Therefore, $d_{MLP} = d_{Mixer} = SC = 2048$.

For all three architectures, the first layer serves as an embedding layer to convert images with 3 RGB channels to $C$ channels and the final layer is a linear classifier.
Following \cite{bachmann2023scaling}, we use Layer Norm for normalization and GELU as the activation function.

\begin{table}[h]
    \vskip 0.15in
    \centering
    \begin{small}
    \begin{tabular}{llll}
        \toprule
        MLP Layer & MLP-Mixer Layer & MLP Output Shape & MLP-Mixer Output Shape \\
        \midrule
        \textbf{Layer-1} & Linear Patch Embedding & [n, 2048] & [n, 16, 128] \\
        \midrule
        \textbf{Layer-2} & Token Mixer MLP 1\dag & [n, 2048] & [n, 128, 16] \\
        \midrule
        \textbf{Layer-3} & Token Mixer MLP 2\dag\dag & [n, 2048] & [n, 16, 128] \\
        \midrule
        \textbf{Layer-4} & Channel Mixer MLP 1 & [n, 2048] & [n, 16, 128] \\
        \midrule
        \textbf{Layer-5} & Channel Mixer MLP 2 & [n, 2048] & [n, 16, 128] \\
        \midrule
      \textbf{Layer-6} & Token Mixer MLP 1\dag & [n, 2048] & [n, 128, 16] \\
        \midrule
        \textbf{Layer-7} & Token Mixer MLP 2\dag\dag & [n, 2048] & [n, 16, 128] \\
        \midrule
        \textbf{Layer-8} & Channel Mixer MLP 1 & [n, 2048] & [n, 16, 128] \\
        \midrule
        \textbf{Layer-9} & Channel Mixer MLP 2 & [n, 2048] & [n, 16, 128] \\
        \midrule
        Linear-10 & Classification Head & [n, 10] & [n, 10] \\
        \bottomrule
    \end{tabular}
    \end{small}
    \caption{Model architectures for MLP (S-MLP and I-MLP) and MLP-Mixer networks used. The fully connected layers with and without weight sharing used for interpolation are bolded. There are two separate Mixer Layers: Layer-2 to Layer-5 and Layer-6 to Layer-9. For both models, the features from the Linear Patch Embedding (Layer-1), the Token Mixer MLP 2 (Layer-3 and Layer 7), Channel Mixer MLP 2 (Layer-5 and Layer-9) are normalized (through a layer norm) while the features between the 2 layers of each Token Mixer (Layer-2, Layer-6) and Channel Mixer (Layer-4, Layer-8) are activated using GELU. The S-MLP and I-MLP models have 39,865,610 parameters, while the MLP-Mixer model uses 93,130 parameters. \\
    \dag indicates that the layer input is transposed \\
    \dag\dag indicates that the output layer is transposed}
    \label{tab:mlp_mixer_model_arch}
\end{table}
\section{Experiment setups} \label{app:experimental-setup}
\paragraph{Setup.} We build on open-sourced PyTorch implementations of S-MLP \cite{bachmann2023scaling}, RepMLPNet \cite{ding2022repmlpnet}, and MLP-Mixer \cite{tolstikhin2021mlpmixer}\cite{wang2022mlpmixerpytorch} and train using the NVIDIA GTX Titan X.
For evaluation, we use $32 \times 32 \times 3$ RGB images from the CIFAR-10 \cite{krizhevsky2009learning} and CIFAR-100 \cite{krizhevsky2009learning} datasets.
We train for 100 epochs with the Adam optimizer using Cross Entropy Loss, a learning rate of 0.0001, and a batch size of 128.
We use S-MLP as our baseline and report the top 1 accuracy score on the test dataset.

\paragraph{Dataset Augmentation.} Similar to Bachmann et al., we note that data augmentation is very important for the S-MLP and I-MLP.
When training for 100 epochs without data augmentation, the S-MLP, I-MLP (for all values of $\alpha)$, and CNN all show signs of overfitting with the validation cross entropy loss starting to increase around 10 epochs.
We also note that the I-MLP with $\alpha=0.5$, I-MLP with $\alpha=1.0$ and CNN perform worse than the S-MLP without data augmentation due to this overfitting issue.
After applying the same data augmentation transforms as Bachmann et al. by using random crops and horizontal flips, we see that accuracy improves for all models and the I-MLP with $\alpha=0.5$ and $\alpha=1.0$ both significantly improve and outperform the S-MLP.

\section{Test-time only interpolation} \label{app:test-time-only-interp}
To validate the correctness of our interpolation experiment, we separately train the MLP and prior models and only use the interpolation weight at test and inference time.
This is equivalent to no interpolation during training, and then interpolating the MLP and prior model once before inference with $\alpha_\text{test}$.
We expect this to perform worse than using the MLP or prior model separately, since the two models will have converged differently.
Our experiments with a CNN prior model shown in Table \ref{tab:nonconstant_interp} confirm that the interpolation at test time performs worse.

\section{Interpolation weight decay: Varying interpolated bias over time} \label{app:interp-weight-decay}
\begin{figure}[!h]
    \centering
    \subfigure[$\alpha$ decay schedule]{
        \includegraphics[width=0.45\textwidth]{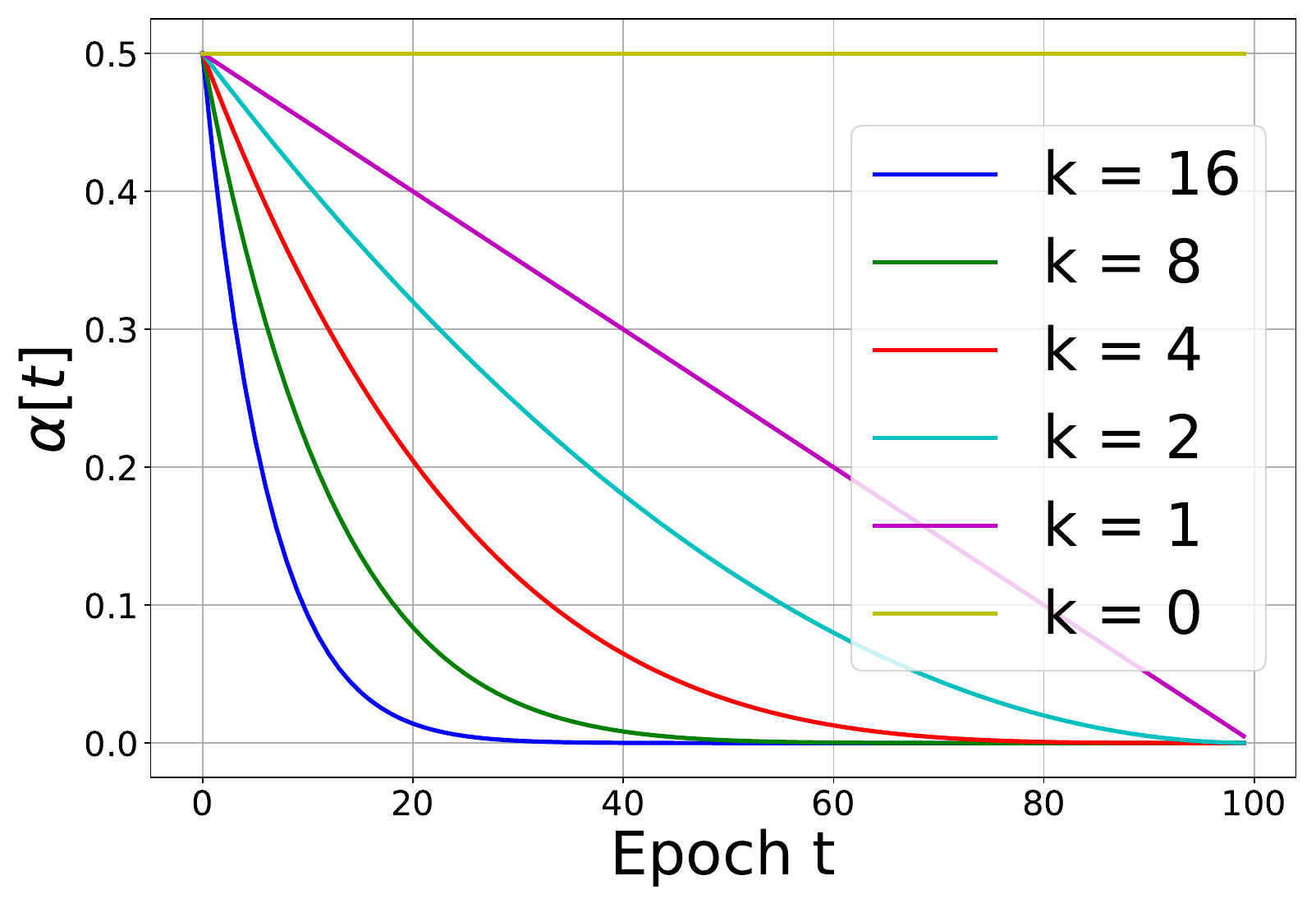}
}
\subfigure[Test accuracy versus $\alpha$ decay rate $k$]{
    \includegraphics[width=0.45\textwidth]{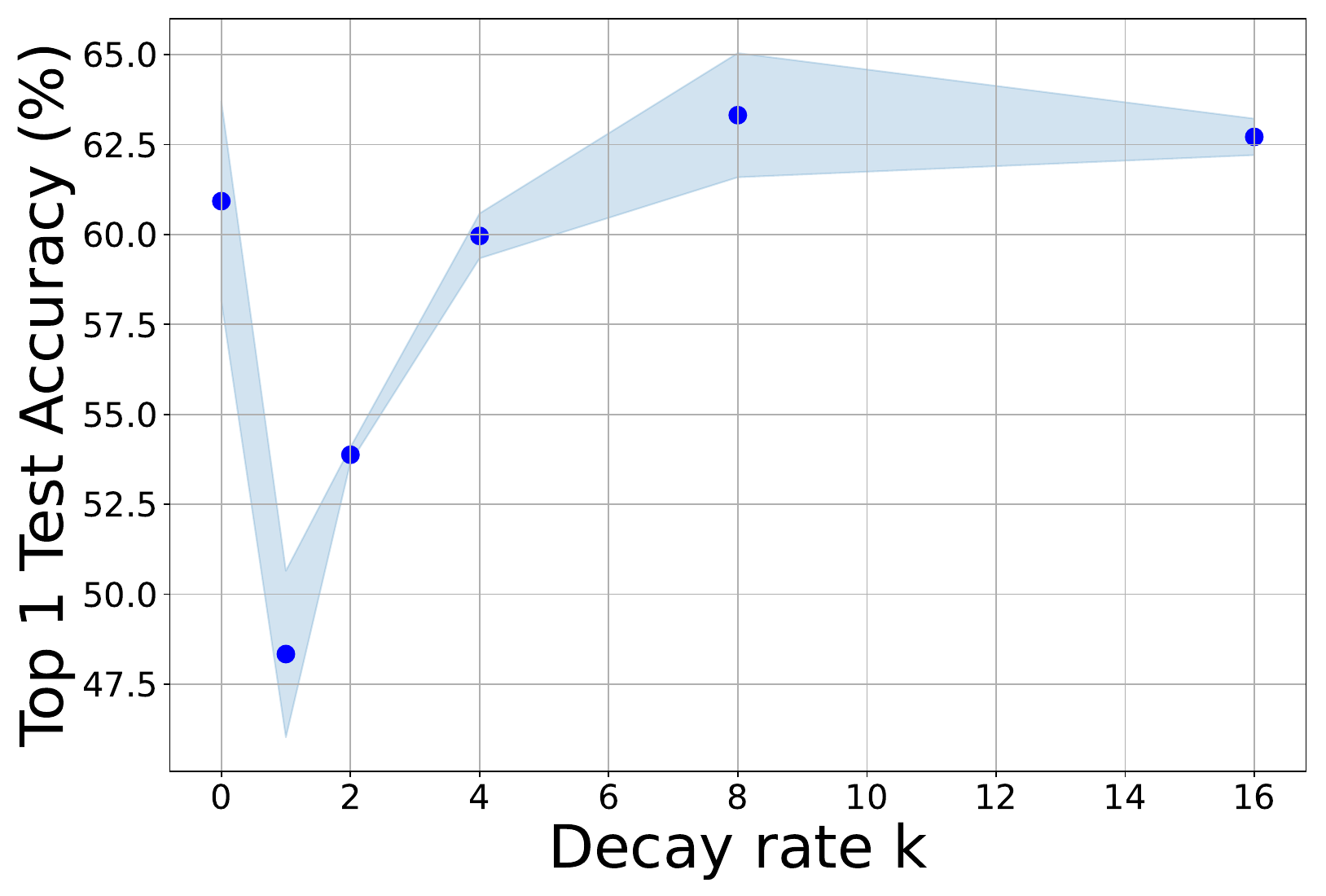}
}
    \caption{CIfAR-10 test accuracy of I-MLP when interpolated with a CNN using varying decay rate $k$ and initial interpolation weight $a=0.5$.}
    \label{fig:interpolation_weight_decay}
\end{figure}

We experimented with non-constant interpolation weights during training with $\alpha[t]$ as a function of the epoch $t$.
We define a decaying interpolation weight schedule of $\alpha[t] = a(1-\frac{t}{t_{max}})^k$, where $a$ represents the initial interpolation weight and $k$ controls the rate of decay.
Note that with our original constant interpolation weight $\alpha[t] = a$, the inductive bias added from the prior model forms a geometric series.
To counteract this compounding effect, we use the decaying weight schedule.
We vary the decay rate $k$ for a fixed initial interpolation weight $a=0.5$ in Figure \ref{fig:interpolation_weight_decay} and Table \ref{tab:nonconstant_interp}.
We observe that constant interpolation with $k=0$ performs similarly to a high decay rate $k>4$, with a sharp local minimum at $k=1$ for linear interpolation weight decay.
We suspect that changing the interpolation weight $\alpha[t]$ leads to each weight update moving in opposite directions and countering the previous weight updates.
With a high decay rate, the interpolation weight $\alpha[t]$ quickly decays to $\alpha[t]=0$ and the weight updates converge in the same direction.

\begin{table}[!h]
    \begin{small}
    \scriptsize
        \begin{tabular}{cc}
            \toprule
            Model & CIFAR-10 \\
            \midrule
            CNN & 62.60 $\pm$ 1.20 \\
            I-MLP no interpolation $\alpha[t]=0$ & 58.36 $\pm$ 0.44 \\
            I-MLP constant interpolation $\alpha[t]=0.5$ & 60.98 $\pm$ 1.97 \\
            I-MLP test-time interpolation $\alpha_\text{test}=0.5$ & 48.63 $\pm$ 1.68 \\
            I-MLP linear decay interpolation $a=0.5, k=1$ & 48.34 $\pm$ 2.31 \\
            I-MLP exponential decay interpolation $a=0.5, k=2$ & 53.88 $\pm$ 0.21 \\
            I-MLP exponential decay interpolation $a=0.5, k=4$ & 59.96 $\pm$ 0.62 \\
            \bottomrule
        \end{tabular}
    \end{small}
    \caption{CIFAR-10 test accuracy scores for a standard MLP, standard CNN, and I-MLP with various types of non-constant inductive bias interpolation methods.}
    \label{tab:nonconstant_interp}
\end{table}

\section{I-MLP in practice: the first layer matters the most} \label{app:first-layer-interp}

We created two separate MLPs (MLP-1 and MLP-2) with a different shape but a similar number of total weight matrix parameters.
Both MLPs are interpolated with a CNN prior model; however, we constrain interpolation to a fixed budget of 944,0256 parameters. 
In MLP-1, the interpolation parameters are spread across 6 layers; 6 layers are interpolated.
In MLP-2, the interpolation parameters are concentrated in the first layer, which is achieved by making the first layer in MLP-2 wider.

To isolate the performance difference to only the interpolation of the first layer versus multiple layers, we confirm that our two MLPs have the a similar number of total parameters and that they perform similarly without interpolation (see baseline in Table \ref{tab:result_comparison}).

We observe that with all interpolation parameters concentrated in the first layer, MLP-2 leverages the inductive bias from CNN-interpolation much more effectively.
The CIFAR-10 top 1 accuracy improved from 54.8\% to 73.7\% in MLP-2 (v.s. 56.4\% to 60.3\% in MLP-1). Similar improvements in CIFAR-100 are also observed.

\begin{table}[!h]
    \centering  
    \begin{small}
        \scriptsize
    \begin{tabular}{ccccccc}
        \toprule
        & & & 
        \multicolumn{2}{c}{CIFAR-10}    &
        \multicolumn{2}{c}{CIFAR-100}    \\ 
        \cmidrule(lr){4-5}
        \cmidrule(lr){6-7}

        Model & \# total params & \# interpolated params & baseline & CNN-interpolated & baseline & CNN-interpolated \\
        \midrule 
        MLP-1 & 16,922,724 & 944,0256 & 56.4\% & 60.3\% & 25.6\% & 27.3\% \\
        \midrule
        MLP-2 & 16,812,024 & 944,0256 & 54.8\% & 73.7\% & 24.4\% & 40.6\% \\
        \bottomrule
    \end{tabular}
    \caption{Parameters and performance comparison of a constant-width (MLP-1) and first-layer-concentrated MLP (MLP-2). Baseline denotes training with no interpolation. CNN-interpolated denotes training with interpolation from a CNN. We show that with a fixed number of interpolated parameters (hence fixed interpolation compute), it is best to concentrate the interpolation parameters in the first layer to capture the most inductive bias.}
    \label{tab:result_comparison}
    \end{small}

\end{table}

\end{document}